\documentclass[]{interact}

\usepackage{epstopdf}
\usepackage[caption=false]{subfig}

\usepackage[numbers,sort&compress]{natbib}
\bibpunct[, ]{[}{]}{,}{n}{,}{,}
\renewcommand\bibfont{\fontsize{10}{12}\selectfont}

\theoremstyle{plain}

\theoremstyle{definition}

\theoremstyle{remark}

\begin{document}

\articletype{Journal of Applied Statistics}

\title{Active Learning-Based Multistage Sequential Decision-Making Model with Application on Common Bile Duct Stone Evaluation}

\author{
\name{Hongzhen Tian\textsuperscript{a}, Reuven Zev Cohen\textsuperscript{b}, Chuck Zhang\textsuperscript{c}\thanks{CONTACT Dr. Hongzhen Tian. Email:hongzhentian@microsoft.com}, Yajun Mei\textsuperscript{c}}
\affil{\textsuperscript{a}WDX DAS, Microsoft, Redmond, WA, US}
\textsuperscript{b}Division of Pediatric Gastroenterology, Hepatology, and Nutrition. Emory University School of Medicine, Children’s Healthcare of Atlanta, Atlanta, GA, US \newline
\textsuperscript{c}H. Milton Stewart School of Industrial \& Systems Engineering, Georgia Institute of Technology, Atlanta, GA, US}

\maketitle

\begin{abstract}

Multistage sequential decision-making scenarios are commonly seen in the healthcare diagnosis process. In this paper, an active learning-based method is developed to actively collect only the necessary patient data in a sequential manner.
There are two novelties in the proposed method. First, unlike the existing ordinal logistic regression model which only models a single stage, we estimate the parameters for all stages together. Second, it is assumed that the coefficients for common features in different stages are kept consistent. The effectiveness of the proposed method is validated in both a simulation study and a real case study. Compared with the baseline method where the data is modeled individually and independently, the proposed method improves the estimation efficiency by 62\%-1838\%. For both simulation and testing cohorts, the proposed method is more effective, stable, interpretable, and computationally efficient on parameter estimation. The proposed method can be easily extended to a variety of scenarios where decision-making can be done sequentially with only necessary information.
\end{abstract}

\begin{keywords}
Active Learning; sequential decision-making; common bile duct stone; common coefficients assumption; incomplete data
\end{keywords}

\section{Introduction}


Multistage sequential decision-making is commonly seen in many real-world applications for the sake of accuracy, efficiency, and cost-effectiveness, such as heathcare audit \cite{ekin2021integrated}, optimal investment \cite{sirbiladze2014multistage}, energy
\cite{gerking1987modeling}, optimization \cite{han2021multiple}, treatment \cite{tao2017adaptive}, and design of experiments \cite{mukkula2020robust}. Meanwhile, active learning is a special case of machine learning in which a learning algorithm can interactively query a user or some other information source to label new data points with the desired outputs \cite{settles2009active}, and has been integrated with decision-making in recent years in various contexts such as imbalanced data \cite{sundin2019active,lee2015Unbalanced} and Bayesian decision-making \cite{filstroff2021targeted}. In this paper, we aim at developing an active learning-based multistage sequential decision-making model to assist doctors making reliable diagnostic decisions and treatment recommendations in a cost-effective and convenient manner.

Our research is motivated from  the evaluation of gallstone. Gallstone are solid particles that can form from cholesterol, bilirubin, and other substances within the gallbladder. These densities are often benign when localized to the gallbladder, but can cause pain, infection, and liver damage when they become stuck in the common bile duct (CBD) and impede the flow of bile into the digestive tract. A stone that becomes impacted in the CBD can be difficult to detect definitively, but its presence requires a procedural intervention, endoscopic retrograde cholangiopancreatography (ERCP). As with most procedures, ERCP requires anesthesia and carries the risk of complications, including pancreatitis, infection, and bleeding. As a result, it is imperative to ensure that ERCP is only performed when there is a stone definitively obstructing the duct. This is particularly true in children \cite{zev2021creation}. 
Thus, one important objective is how to efficiently predict the presence of a CBD stone in a child with high accuracy and specificity. 



From the machine learning or statistical point of view, this might be straightforward:  one would like to collect as many informative features as possible for each patient, which will lead to higher prediction accuracy. However, from the medical point of view, this is highly non-trivial, since some informative features could be expensive and hard to obtain in many real-world scenarios.
Obtaining additional clinical data often comes with a price, including delayed time to intervention, cost, and complications. This price can be particularly high in pediatric care. For instance, while magnetic resonance imaging (MRI) is the most accurate way to diagnose a CBD stone, only 5\% of  pediatric  patients in our real dataset have such diagnostic, as pediatric MRI often involves sedation, long scheduling time, etc. Indeed a less cost-effective approach such as laboratory tests or ultrasound (US) often might be sufficient. Thus a common clinical challenge is how to reduce the cost of data collection for diagnosis decision-making, yet avoid computational issues associated with limited or incomplete data. 

In this article, a multistage sequential decision-making model is developed to assist doctors making reliable diagnostic decisions and treatment recommendations with high accuracy in a cost-effective and convenient manner. In our approach, some patients are allowed to take fewer examinations if their symptoms are apparent according to early-stage features, and some patients will be recommended to take more examinations if their symptoms are ambiguous and doctors have difficulty making reliable treatment recommendations.

There are several benefits to the proposed model.
This novel platform allows the healthcare provider to actively collect only the necessary data sequentially rather than collecting all the data at inception. As a result, the diagnostic process can be faster, more convenient, and cost effective, while retaining significant accuracy.
In the research setting, our model offers a novel solution to address challenges raised by incomplete data through offering sequential decision-making in an active learning way.
The active learning-based model allows some missing data yet guarantees reliability and prediction accuracy, which makes the incomplete data missing ``at control" rather than missing at random.

In the proposed framework, one important thing is to identify a proper predictive model for each stage as the building block.
General categorization is required to utilize the model. Patients must be stratified to: certainly healthy (``0"), indeterminate (``0.5"), and certainly sick (``1").
After comparing the performance of different machine learning models, including neural networks, random forest, logistic regression, we selected the logistic regression model, which provides stability and accuracy with easy interpretability, see our previous work \cite{zev2021creation}.
In the sections to follow, a generalized logistic regression model will be utilized as a building block for each stage.

Multiclass logistic regression, also called  softmax/multinoulli/multinomial regression, is a classification method that generalizes logistic regression to multiclass problems ($K>2$). The idea behind this classification model is one-vs-all or one-vs-rest, which means $K$ binary classifier models will be trained to ensure every class has a binary classifier to identify whether a sample belongs to this specific class or not.
This model enables multiple outcomes but does not involve the ordering of the categorical dependent variable.
The ordinal logistic regression model, also called the ordered logit model or proportional odds model, overcomes this limitation by calculating the log of the odds of cumulative events \cite{mccullagh1980regression,kim2004topics,watson1986generalized}.
Unlike simple logistic regression, ordinal logistic models consider the probability of an event and all the events that are below the focal event in the ordered hierarchy.
The existing ordinal logistic regression model satisfies our needs in single stages but failed to deal with multistage scenarios simultaneously.
Therefore, in this research, we'll develop our own multistage model with the existing ordinal logistic regression model as a building block for each stage, but estimate the parameters in all stages together.

The contribution of our work is twofold. In terms of statistical analysis, an active learning model is proposed in a multistage decision-making process. This learning algorithm can interactively query a patient collect additional clinical data for collecting necessary information \cite{olsson2009literature}.
Also, the proposed model introduces a novel way of dealing with incomplete data by allowing some data to be absent and modeling with incomplete but necessary data directly, rather than guessing the incomplete data with uncertainty.
In terms of healthcare application,
a general strategy is developed to help the doctor making diagnostic decisions and treatment recommendations sequentially when some laboratory data are more expensive, invasive, or time-consuming to collect than other data.
By recommending requisite testing to those patients in whom symptoms do not support a definitive diagnosis, the proposed process is fast, cost-effective, yet reliable. 
Also, the methodology applies to a large variety of disease diagnose that involve a sequence of indirect examinations and some unpleasant or inaccurate examinations.

Note that our proposed model/method is closely related to incomplete data analysis in the statistics literature, where many widely-used methods have been developed, e.g., simple deletion on feature or record, mean/median/mode/zeros substitution, regression imputation, last observation carried forward, EM, etc. \cite{kang2013prevention,ma2016missing}. However, these existing methods were developed from two primary approaches. One is deleting the records or features with missing values, which will lead to a waste of available data and some intrinsic problems associated with limited data size. The other approach is ``guessing" or ``learning" the missing data from the observed data, which can be dangerous for clinical modeling since the synthetic data brings extra uncertainties. Our approach is different from these existing methods in the sense that we actively decide which kinds of data to be observed and which kinds of data to be missing.

In Section 2, the proposed model is described. 
Section 3 discusses the novel approach to parameter estimation for the proposed model.
In Section 4, the performance of the proposed methodology on synthetic data is reported. Finally, in section 5, the analytical result on real clinical data is demonstrated.

\section{Proposed Active Learning Multistage Sequential Decision-Making Model}
In this section, 
we first present a high-level framework of our proposed active learning-based multistage sequential decision-making model, and then introduce necessary notation to represent observed data in our framework. Finally, our model is rigorously developed based on the existing ordinal logistic regression model.


\subsection{Model Framework}


\begin{figure}[t]
   \centering
   \includegraphics[width = \textwidth]{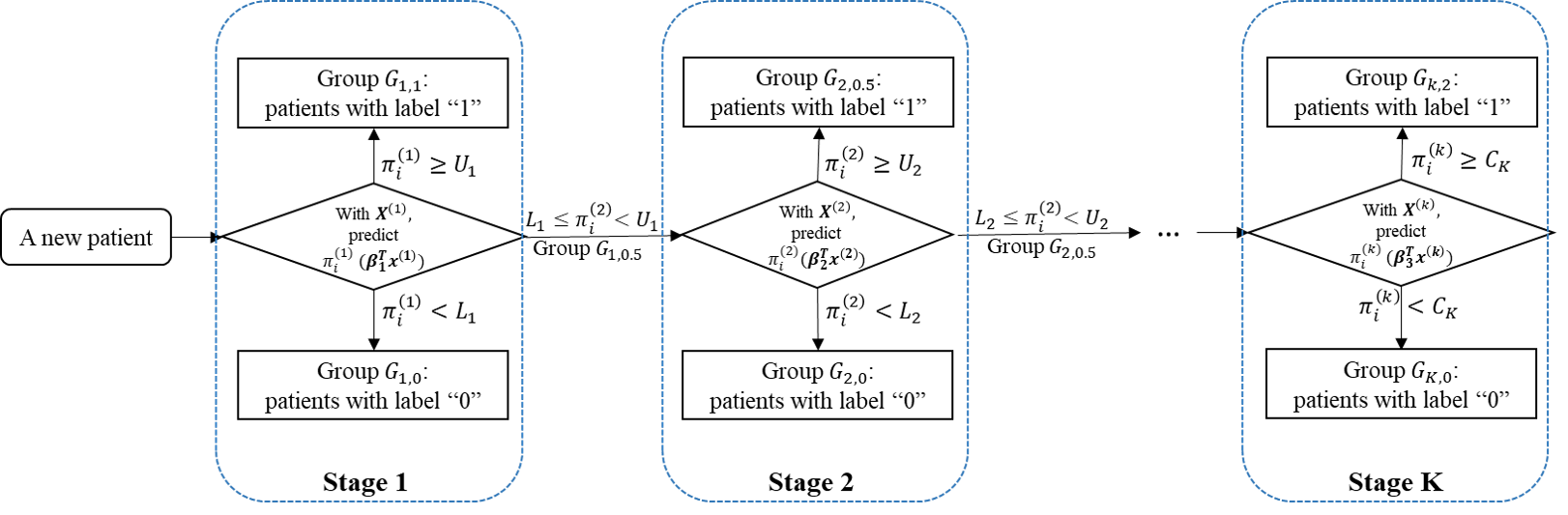}
    \caption{The framework of the proposed multistage sequential decision-making model, where $K$ is the number of stages, ${\pi}_i^{(j)}$ is the probability of patient $i$ having disease estimated in stage $j$, $U_j$ and $L_j$ is the cutoff points on the estimated probability in stage $j$, and $C_{K}$ is the single cutoff point in the last stage, $K$.}
   \label{fig_framework3}
\end{figure}

The proposed multistage sequential decision-making model is shown in Fig~\ref{fig_framework3}. To achieve the goal of optimizing the diagnostic process, under our proposed model, the features of a patient's clinical data are classified into several different stages according to the relevant cost, invasiveness, time, patient's willingness to participate, etc. The earlier stages involve features that are patient-friendly, inexpensive, though potentially less informative. The latter stages could involve some more time-consuming, expensive, unpleasant, but potentially more informative examinations. As in group sequential clinical trials, if we can make a definitive yes or no decision at a given stage, then we will stop taking features/observations and make the corresponding decision; but if we are not sure, then we will move to the next stages by taking additional features/observations from the patient.

To be more specific, for the $k$-th stage, based on the cumulative features, an ordinal logistic regression model is trained as the classifier to estimate the probability of sickness for each patient ($\pi_{i}^{(k)}$). For every classifier, the corresponding optimal cutoff points, $L_k, U_k$, or $C_K$ for the last stage, will be augment clinical decision making: if $\pi_i^{(k)} < L_k$, patient $i$ will be labeled as healthy (``0") and will not undergo further intervention; if $\pi_i^{(1)} > U_k$, patient $i$ will be labeled as sick (``1"), they will proceed to the recommended intervention, such as ERCP or surgery; otherwise,  the patient $i$ will be labeled as indeterminate (``0.5") and undergo further diagnostic workup. 
Following the last stage, as no further diagnostic examinations are available, the provider must make a treatment recommendations based on a single threshold, $C_{K}$, and the patient will be classified as either healthy or sick.
By quantifying the probability of being sick via predictive models based on currently available features, the diagnosis could be more objective and stable compared with an empirical estimation.

In the evaluation of a newly presenting patient,
this framework with trained parameters can be utilized to arrive at a diagnosis and recommend treatments accordingly. In such a way, the necessary features will be collected sequentially and actively.

In the proposed multistage model, ordinal logistic regression is adopted as the basic predictive model in each stage. A novel feature of this model is, the patients' groups in the different stages have a nested structure rather than independent. Further, the features in the current stage $k$ include all those of the former stages and newly added features in the given category $k$. 
This allows us to consider the predictive models in different stages together rather than in an isolated fashion. The details will be discussed further in the parameter estimation discussion in Section 3.


\subsection{Data Organization for 2-stage Model as Example}
To illustrate the proposed active learning style multistage sequential decision-making model, the 2-stage decision-making model will be elaborated as an illustrative example.
As discussed earlier, all the capturable features are first divided into several categories, each corresponds to a certain stage in our multistage model.
The criteria for feature grouping are flexible and can be adjusted according to real scenarios case-by-case: 
1) For the features inside a category, the cost of collection and the information contained therein should be similar. For example, all indices can be obtained from one examination such as a blood test.
   2) Each category should contain some useful information, and excess categories should be avoided. For example, there is no need to treat each variable as a separate category,  
   especially less informative variables.

To illustrate the data organization for the proposed model, without loss of generality, we can take the 2-stage sequential decision-making model 
as an example as follows.
In the observed (training) data, denote by $N_1$ and $N_2$  the number of patients in  stage 1 and stage 2, respectively, and denote by $p_1$ and $p_2$ the number of features collected in the two stages. In the first stage, the observed features or explanatory variables  are denoted by $\boldsymbol{X}^{(1)}= [x_{1}^{(1)}, x_{2}^{(1)}, \ldots, x_{p_{1}}^{(1)}]$ and the corresponding decision variable.  $Y^{(1)} \in \{``0", ``0.5", ``1"\}$, where $Y_{i}^{(1)} = ``0.5"$ implies that the more features need to be collected on patient $i$ in stage 2 for diagnosis. 
In stage 2,  the new observed features or explanatory variables  are denoted by $(x_{1}^{(2)}, x_{2}^{(2)}, \ldots, x_{p_{2}}^{(2)})$ and the corresponding decision variable $Y^{(2)} \in \{``0", ``1"\}.$ Note that the feature set from stage 1 will be used in stage 2 cumulatively, and thus the number of features in stage 2 is actually $p_{1}+p_{2}$.

Hence, for the 2-stage model, the (training) data can be summarized as:
\begin{itemize}
    \item Stage 1: $\left\{Y_{i}^{(1)}; x_{i,1}^{(1)}, x_{i,2}^{(1)}, \ldots, x_{i, p_{1}}^{(1)}\right\} =\left\{Y_{i}^{(1)}; \boldsymbol{X}_{i}^{(1)}\right\}, i=1,2, \ldots, N_{1}, \\
    Y_{i}^{(1)} \in \{{``0", ``0.5", ``1"}\};$

    \item Stage 2: $\left\{Y_{i}^{(2)}; x_{i,1}^{(1)}, x_{i,2}^{(1)}, \ldots, x_{i,p_{1}}^{(1)}; x_{i,1}^{(2)}, x_{i,2}^{(2)}, \ldots, x_{i,p_{2}}^{(2)}\right\}=\left\{\boldsymbol{Y}^{(2)}; \boldsymbol{X}^{(2)}\right\}, i=1,2, \ldots, N_{2},
    Y_{i}^{(2)} \in \{{``0", ``1"}\};$
\end{itemize}

For a new patient, after observing $\boldsymbol{X}_{new}^{(1)} = [x_{new,1}^{(1)}, x_{new,2}^{(1)}, \ldots, x_{new, p_{1}}^{(1)}],$ we need to predict $Y_{new}^{(1)} \in \{{``0", ``0.5", ``1"}\}.$ When we predict $Y_{new}^{(1)} \in \{``0", ``1"\},$ this is the final decision. When we predict  $Y_{new}^{(1)} = ``0.5",$ we will then take more observations from this patient in the stage 2, i..e, $[x_{new,1}^{(2)}, x_{new,2}^{(2)}, \ldots, x_{new, p_{2}}^{(2)}],$ and make another prediction  $Y_{new}^{(2)} \in \{``0", ``1"\},$ which will be our final decision.


\subsection{Our Proposed Model} 

In our proposed model, the building block is an ordinal logistic regression model, which can serve as a filter in each stage to split patients into 3 groups based on the ordered probability of being sick. For better presentation, let us first provide a quick review of ordinal logistic regression, which can be seen a form of a generalized linear model (GLM) and an extension of binary logistic regression.  
Under ordinal logistic regression model, we assume that there is an underlying continuous latent variable
\begin{equation}
Y^{*}_{i}=\boldsymbol{\beta}^\top\boldsymbol{x}_{i}+\boldsymbol{\varepsilon}_{i},
\label{eq:ystar}
\end{equation}
where the error term $\boldsymbol{\varepsilon}_{i}$ has a logistic distribution with the cumulative distribution function (CDF) $F(t) = P( \boldsymbol{\varepsilon}_{i} \le t) = \frac{1}{1+e^{-t}}.$ Of course, other distributions of  $\boldsymbol{\varepsilon}_{i}$'s are possible such as Gaussian/normal distribution. Then the observable response variable $Y_i$ will fall into several ordinal categories based on the values of latent variable $Y^{*}_{i}$ and the cutoff points.

%

Now we are ready to present our proposed 2-stage sequential decision-making model. We assume that there are two latent variables, $Y^{(1)*}$ and  $Y^{(2)*}$ for the two stages as defined in (\ref{eq:ystar}). Then the decision at each stage is given by
\begin{equation}
Y_{i}^{(1)}=\left\{\begin{array}{ll}
0, & \text { if } Y^{(1)*}_i<L_{1}^{*} \\
0.5, & \text { if } L_{1}^{*} \leq Y^{(1)*}_i<U_{1}^{*} \\
1, & \text { if } Y^{(1)*}_i \geq U_{1}^{*}
\end{array}\right.
,
\quad
Y_{i}^{(2)}=\left\{\begin{array}{ll}
0, & \text { if }  Y^{(2)*}_i <C_{2}^{*} \\
1, & \text { if }  Y^{(2)*}_i \geq C_{2}^{*}
\label{eq:our2model}
\end{array}\right.\end{equation}
Note that the cutoff points $\left\{ L_{1}^{*}, U_{1}^{*}, C_{2}^{*} \right\}$ applying on the latent variable $Y^{*}$. This is equivalent to applying the cutoff points $\left\{ L_{1}, U_{1}, C_{2} \right\}$ on the probability estimation $\pi_i^{(k)}$ in the Fig~\ref{fig_framework3} after going through a link function transformation.

Recall 
that in our proposed multistage model, the output $Y_{i}^{(1)} \in \{{``0", ``0.5", ``1"}\}$ at the first stage represents healthy, indeterminate, and sick, respectively. At the last stage,  $Y_{i}^{(2)} \in \{{``0", ``1"}\}$ indicates healthy and sick. It is useful to mention that the outputs are ordered, i.e. $Y_i^{(1)} \leq Y_{i}^{(2)}.$

A subtlety in our proposed multistage model is the relationship of the parameter $\boldsymbol{\beta}$ in (\ref{eq:ystar}) across different stages.
A computationally simple baseline approach is to assume that they can be completely different, where we can simply apply ordinal logistic regression models to estimate parameters each stage individually and independently. Unfortunately, this does not consider the nested structure and potential correlations between the two adjacent stages, and thus might cause mis-understanding and confusing interpretation in practice. In our paper, we will make a further assumption that $\boldsymbol{\beta}^{(1)}$ and $\boldsymbol{\beta}^{(2)}$ enjoys the nested structure in the sense of having the same $\beta$ values with the same features, i.e.,
\[
\boldsymbol{\beta}^{(2)} =\left\{ \beta^{(1)}_1,\beta^{(1)}_2, \ldots, \beta^{(1)}_{p_1}; \beta^{(2)}_1, \beta^{(2)}_2, \ldots, \beta^{(2)}_{p_2} \right\}= \left\{ \boldsymbol{\beta}^{(1)}; \beta^{(2)}_1, \beta^{(2)}_2, \ldots, \beta^{(2)}_{p_2} \right\}.
\]
Under this common coefficients assumption, the newly added features from additional stages will not affect the weights of existing ones, which makes the model more interpretable for doctors and patients.

\section{Parameter Estimation}

At the high-level, we propose to use the maximum likelihood estimation (MLE) method to estimate not only the unknown $\beta$'s values in different stages, but also the cutoff values $\left\{ L_{1}^{*}, U_{1}^{*}, C_{2}^{*} \right\}$. Note that the coefficients $\beta$'s  can also be estimated by the penalized likelihood estimation framework such as LASSO for variable selection when there are a large of available features, which is beyond the scope of this article and will be presented elsewhere. Here we assume that we pre-screen all features, and only use those useful features in our model and medical decision making.



To better present our main ideas, let us begin with a single stage with a ordinal logistic regression model. To simplify the notation, we assume that the response $Y$ results from the underlying latent variable $Y^*$ by using cutoff points $\theta_{1} < \theta_{2}$:
\begin{equation}
Y=\left\{\begin{array}{lr}
0, \text { if }  \theta_{0} < Y^{*}  \leq \theta_{1} \\
0.5, \text { if } \theta_{1}<Y^{*} \leq \theta_{2} \\
1, \text { if } \theta_{2}<Y^{*} \leq \theta_{3} \\
\end{array}\right. ,
\label{eq:olr}
\end{equation}
where $\theta_0 = -\infty$ and $\theta_3 = \infty$ are introduced for the technical reason to simplify the notation.  Also note that at the last stage, we have $\theta_{1} = \theta_{2}$, which will force one to make a decision $\{Y=0\}$ or $\{Y=1\}.$ Under this notation,  the conditional distribution of $Y$ is given by
\begin{equation}
\begin{aligned}
P(Y=j \mid \boldsymbol{x}) &=P\left(\theta_{j-1}<Y^{*} \leq \theta_{j} \mid \boldsymbol{x}\right)
=P\left(\theta_{j-1}<\boldsymbol{\beta}^{\top}  \boldsymbol{x}+\varepsilon \leq \theta_{j}\right) \\
&=F\left(\theta_{j}-\boldsymbol{\beta}^{\top}  \boldsymbol{x}\right)-F\left(\theta_{j-1}-\boldsymbol{\beta}^{\top}  \boldsymbol{x}\right)
\end{aligned}
\label{eq:probn}
\end{equation}
where $\boldsymbol{\beta}, \theta_{j}$ are model coefficients and cutoff points to be estimated from the data, $P$ is the abbreviation of probability, and $F$ is the CDF of the error term $\boldsymbol{\varepsilon}$ which takes on the role of the inverse link function.

For a given stage or a given ordinal logistic regression model with $N$ subjects, the joint log-likelihood function becomes
\begin{equation}
\log \mathcal{L}\left(\boldsymbol{\beta}, \boldsymbol{\theta} \mid \boldsymbol{X}, Y\right)=
\sum_{i=1}^{N}
\sum_{j=1}^{J} I(y_{i}=j) \log \left[ F\left(\theta_{j}-\boldsymbol{\beta^{\top}} \boldsymbol{x}_{i}\right)-F\left(\theta_{j-1}-\boldsymbol{\beta^{\top}} \boldsymbol{x}_{i}\right) \right]
\notag
\end{equation}
It has been well investigated to maximize this log-likelihood objective function which possesses several nice properties.
McCullagh presented a Fisher scoring algorithm for ML estimation, expressing the likelihood function using cumulative probabilities. McCullagh showed that a sufficiently large $n$ guarantees a unique maximum of the likelihood \cite{mccullagh1980regression,agresti2003categorical}.
Burridge and Pratt showed that the log-likelihood is concave for many cumulative link models, including the logit, probit, and complementary log-log \cite{agresti2003categorical}. Iterative algorithms usually converge rapidly to the ML estimates.
The optimization computing algorithms are developed based on these useful properties. The existing approaches compute the MLE by iteratively reweighted least squares (IRLS). Actually, for packages in $R$ software, all GLM are fit using the IRLS  estimation method for the log-likelihood.


Now let us go back to the parameter estimation of our proposed multistage model. For simplicity,  denote $[\boldsymbol{\beta}, \boldsymbol{\theta}]$ as $\boldsymbol{\zeta}$, and $\log \mathcal{L}\left( \boldsymbol{\zeta} \mid \boldsymbol{X}, Y\right)$ as $\ell(\boldsymbol{\zeta})$.
Under our notation, the joint log-likelihood function for all samples in all stages will be as follows:
\begin{equation}
\begin{aligned}
\ell(\boldsymbol{\zeta})=
\sum_{k=1}^{K}
\sum_{i=1}^{N_k}
\sum_{j=1}^{J_{k}} I(y_{i}^{(k)}=j) \log \left[ F\left(\theta^{(k)}_{j}-\boldsymbol{\beta}^{(k)\top} \boldsymbol{x}_{i}^{(k)}\right)-F\left(\theta^{(k)}_{j-1}-\boldsymbol{\beta}^{(k)\top} \boldsymbol{x}_{i}^{(k)}\right) \right]
\notag
\end{aligned}
\end{equation}
where 
$K$ is the number of stages in the proposed model, $N_{k}$ is the number of patients in the $k^{th}$ stage, and $J_k$ is the number of categories in the $k^{th}$ stage for which we have
\begin{equation}
J_{k}=\left\{\begin{array}{ll}
3, \text { if } 1 \leq k < K \\
2, \text { if } k=K
\end{array}\right.
\notag
\end{equation}




Given the joint likelihood function $\ell(\boldsymbol{\zeta})$, the next step is to simultaneously estimate $\boldsymbol{\zeta}$, all the parameters in the $K$ stages, that maximizes the likelihood function. 
Here we propose to adopt the gradient descent algorithm and a key step is to compute the gradient for each parameter, which can be written as:
\begin{equation}
\frac{\partial \ell(\boldsymbol{\zeta})} { \partial \boldsymbol{\beta}^{(k)}}=
\sum_{i=1}^{N_{k}}
\sum_{j=1}^{J_{k}} I(y_{i}^{(k)}=j) \frac{ (-\boldsymbol{x}_{i}^{(k)}) \left[ f\left(\theta_{j}^{(k)}-\boldsymbol{\beta}^{(k)\top} \boldsymbol{x}_{i}^{(k)}\right) - f\left(\theta_{j-1}^{(k)}-\boldsymbol{\beta}^{(k)\top} \boldsymbol{x}_{i}^{(k)}\right) \right] }
{  F\left(\theta_{j}^{(k)}-\boldsymbol{\beta}^{(k)\top} \boldsymbol{x}_{i}^{(k)}\right)-F\left(\theta_{j-1}^{(k)}-\boldsymbol{\beta}^{(k)\top} \boldsymbol{x}_{i}^{(k)}\right) }
\notag
\end{equation}
\begin{equation}
\begin{aligned}
\frac{\partial \ell(\boldsymbol{\zeta})} { \partial \theta_{j}^{(k)}}=
\sum_{i=1}^{N_{k}}
\sum_{j=1}^{J_{k}} I(y_{i}^{(k)}=j)&\{
\frac{ f\left(\theta_{j}^{(k)}-\boldsymbol{\beta}^{(k)\top} \boldsymbol{x}_{i}^{(k)}\right)}
{  F\left(\theta_{j}^{(k)}-\boldsymbol{\beta}^{(k)\top} \boldsymbol{x}_{i}^{(k)}\right)-F\left(\theta_{j-1}^{(k)}-\boldsymbol{\beta}^{(k)\top} \boldsymbol{x}_{i}^{(k)}\right) }  \\
&-\frac{ f\left(\theta_{j}^{(k)}-\boldsymbol{\beta}^{(k)\top} \boldsymbol{x}_{i}^{(k)}\right)}
{  F\left(\theta_{j+1}^{(k)}-\boldsymbol{\beta}^{(k)\top} \boldsymbol{x}_{i}^{(k)}\right)-F\left(\theta_{j}^{(k)}-\boldsymbol{\beta}^{(k)\top} \boldsymbol{x}_{i}^{(k)}\right)  }
\}
\end{aligned}
\notag
\end{equation}
where 
$f(t) = F'(t)$ is the probability density function (PDF) of  $\boldsymbol{\varepsilon}$:
 $
 f(t) = \frac{1}{(e^{t/2} + e^{-t/2})^2}$.

We should also remark some technical details in computation.
Note that the coefficient in stage $k$, $\boldsymbol{\beta}^{(k)}$, contains $\boldsymbol{\beta}^{(k-1)}$ and newly added coefficients for features in the $k^{th}$ stage. This makes the coefficients in different stages of different lengths. To address this issue, we propose finding the gradient in each stage $\frac{\partial \ell(\boldsymbol{\zeta})} { \partial \boldsymbol{\beta}^{(k)}}$ individually first. Then, adding them together with the front ends aligned and pad the shorter ends with 0's:
$\frac{\partial \ell(\boldsymbol{\zeta})} { \partial \boldsymbol{\beta}} = \left[\frac{\partial \ell(\boldsymbol{\zeta})} { \partial \boldsymbol{\beta}^{(1)}}; 0,0,...,0 \right] + ... + \left[ \frac{\partial \ell(\boldsymbol{\zeta})} { \partial \boldsymbol{\beta}^{(K)}}\right]
$.
In addition, the indicator function $I(y_i^{(j)} = j)$ can be replaced with a mask matrix with boolean values. Alternatively, the indicator function can be avoided by data preprocessing. 
For instance, we can split the patients into different groups to mark the identity, e.g., using $G_{k,j}$ to denote the group of patients who have label $j$ in stage $k$, see \textit{Appendix A} for more details.

To numerically solve our proposed joint-optimization problem, we follow the framework of the existing $polr()$ function in the package MASS in R software, and adopt the $optim()$ function as optimization solver. 
The $optim()$ is a function designed for general-purpose optimization based on Nelder--Mead, quasi-Newton and conjugate-gradient algorithms. 
In $optim()$, the Broyden–Fletcher–Goldfarb–Shanno algorithm (BFGS) method, an iterative method for solving unconstrained nonlinear optimization problems \cite{avriel2003nonlinear}, is adopted, since  it works best with analytic gradients based on our extensive experiences.

\section{Simulation Study}
In this section, a simulation is conducted to verify the effectiveness and efficiency of the proposed methodology.
First, a 2-stage synthetic dataset, $\{X^{(1)}, Y^{(1)}; X^{(2)}, Y^{(2)} \},$ is generated with prescribed coefficients $\boldsymbol{\beta}^{(1)}, \boldsymbol{\beta}^{(2)}$ and classification thresholds $\{ L^{*}_1, U^{*}_1, C^{*}_2\}$.
Then, 
we conduct both a baseline approach 
and the proposed approach on the synthetic dataset 
to demonstrate the performance of the proposed method in the aspect of prediction accuracy, stability, interpretability, and relative efficiency.

Let $N_{1}=10,000$ denote the total number of synthetic patients at the outset.
We design the features of patients as follows:
$X_1 \sim Bernoulli (0.3)$, $X_2 \sim N (-1, 1)$, $X_3 \sim N (1, 1)$, $X_4 \sim N (0, 2)$, $X_5 \sim Bernoulli (0.4)$, $X_6 \sim N (-1, 1)$, $X_7 \sim N (0, 1)$.
For stage 1, we design feature set as: $\boldsymbol{X}^{(1)}=\left\{ X_1, X_2, X_3, X_4 \right\}$.
For stage 2, we add 3 new features: $\boldsymbol{X}^{(2)}= \left\{ X_1, X_2, X_3, X_4; X_5, X_6, X_7 \right\}= \left\{\boldsymbol{X}^{(1)}; X_6, X_7, X_8 \right\}$.
The error term $\boldsymbol{\varepsilon}$ follows standard logistic regression for both stages:
$\boldsymbol{\varepsilon^{(1)}} \sim Logistic(0, 1), \quad \boldsymbol{\varepsilon^{(2)}} \sim Logistic(0, 1)$. 
Note that if the error term $\boldsymbol{\varepsilon}$ follows logistic regression with other parameters, 
the estimation of $\boldsymbol{\beta}$ and $\boldsymbol{\theta}$ will be scaled, 
see Eq.~(\ref{eq:our2model}). 
The coefficients are designed as: $\boldsymbol{\beta}^{(1)} = [2, 2, 2, 2], \quad \boldsymbol{\beta}^{(2)} = [\boldsymbol{\beta}^{(1)}, 4, 4, 4] = [2, 2, 2, 2, 4, 4, 4] $
which satisfys the common coefficients assumption. 
Next, we can calculate the latent variable $Y^*$ according to Eq.~(\ref{eq:ystar}). 
To cut the $Y^*$ into ordered categories, we design the cutoff points as $\boldsymbol{\theta}= \{ L_1^{*}, U_1^{*}, C_2^{*} \} = \{-2.2, 2.2,  0.5\}$.
The value $-2.2$ and $2.2$ are chosen since $U_1 = F(2.2) = 0.9$ and $L_1 = F(-2.2) = 0.1$, where $0.9$ and $0.1$ are the designed cutoff points on probability $Y$. Via such design, 
we have $N_{2}  \approx \frac{1}{3}N_{1}$. 
Given $\boldsymbol{\theta}$, the corresponding true label $Y$ can be obtained according to Eq.~(\ref{eq:our2model}). 
In the baseline approach, the 2 stages proceed independently: in stage 1, a standard ordinal logistic regression is applied on $\{\boldsymbol{Y}^{(1)},\boldsymbol{X}^{(1)}\}$ via $polr()$ function in R; and in stage 2, a standard logistic regression is applied on $\{\boldsymbol{Y}^{(2)},\boldsymbol{X}^{(2)}\}$ via the $glm()$ function in R.
While in the proposed method, all the parameters in 2 stages are estimated together via MLE.
To provide reliable statistical inference, we adopt the Monte-Carlo method and run the data generation and parameter estimation 100 times.

\begin{figure}[t]
    \centering
    \includegraphics[width =0.7\linewidth]{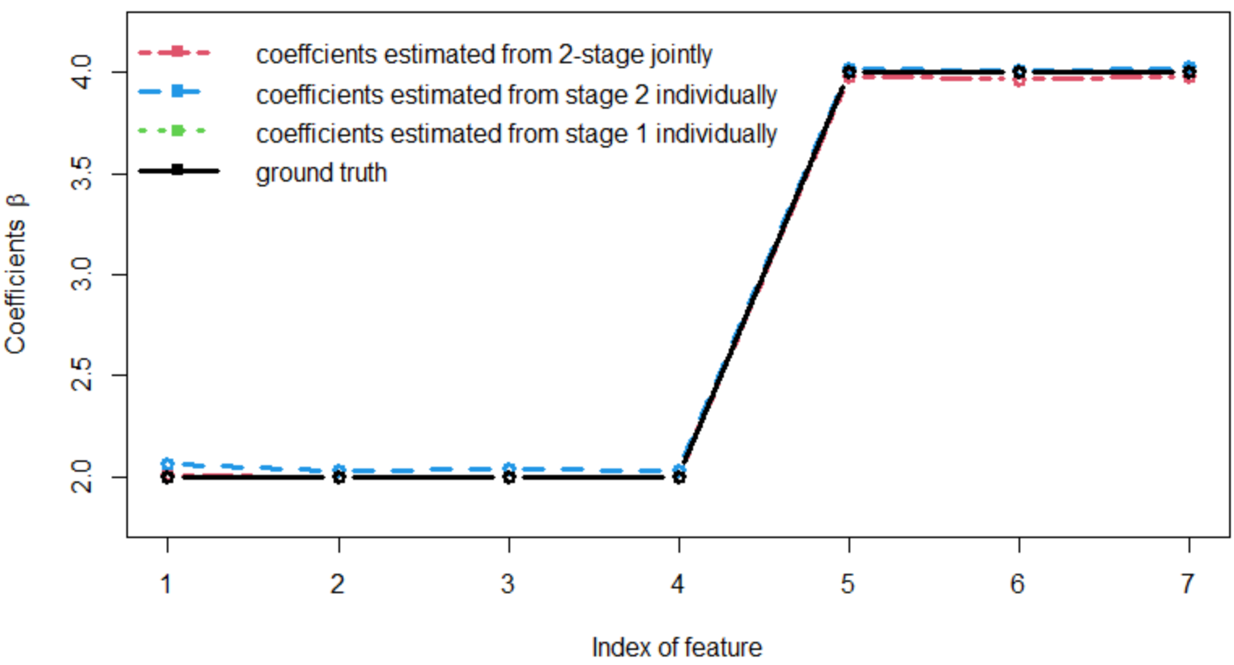}
    \caption{Mean coefficients estimation comparison among ground truth (black line), mean coefficients estimated from baseline approached individually (blue and green line), and mean coefficients estimated from proposed method jointly (red line). This plot demonstrates that both methods have little or no bias when estimation parameters, proving the effectiveness of both methods on estimating.}
    \label{fig:coefsimu}
\end{figure}

The comparison of the mean coefficients estimated from the two methods is visualized in Fig. ~\ref{fig:coefsimu}. Clearly, the mean coefficients estimated from the two methods are similar and both coincide well with ground truth, indicating that both methods have little or no bias on mean parameter estimation, proving the effectiveness of both methods on estimating.
To quantify the subtle difference, 
we can utilize  mean square error (MSE) to measure the distance between the estimated coefficients and the ground truth. 
For coefficients in stage 1, the MSE of the estimation from the baseline method and the proposed method  are $7.62 \times 10^{-6}$ and $2.25 \times 10^{-6}$, respectively. Meanwhile, for stage 2, the MSE of the baseline method and the proposed method are 0.0011 and 0.0003, respectively. This indicates a slight advantage of the proposed methods on mean coefficients estimation. See Appendix B for more comparison on mean prediction metrics.

\begin{figure}[t]
\centering
\begin{minipage}[t]{0.48\linewidth}
\centering
\includegraphics[width=6cm]{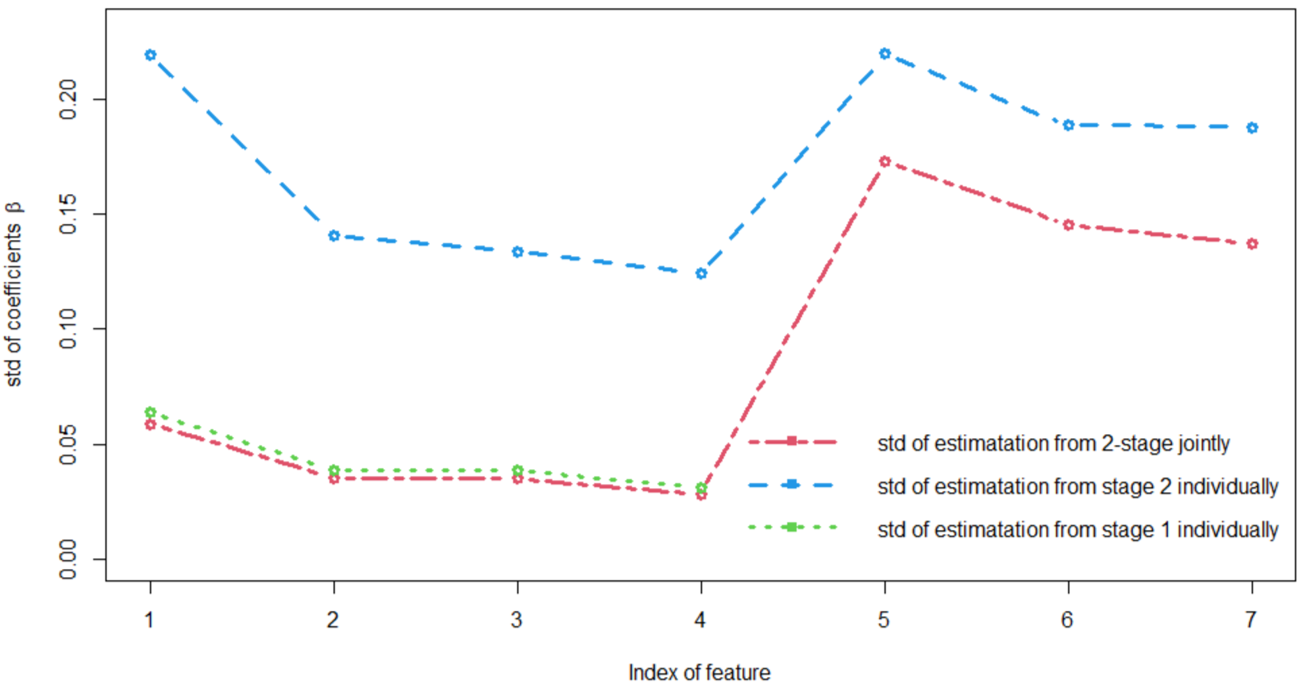}
\caption{Standard deviation of coefficients estimation from two methods. }
 \label{fig:stdcoefsimu}
\end{minipage}
\begin{minipage}[t]{0.48\linewidth}
\centering
\includegraphics[width=6cm]{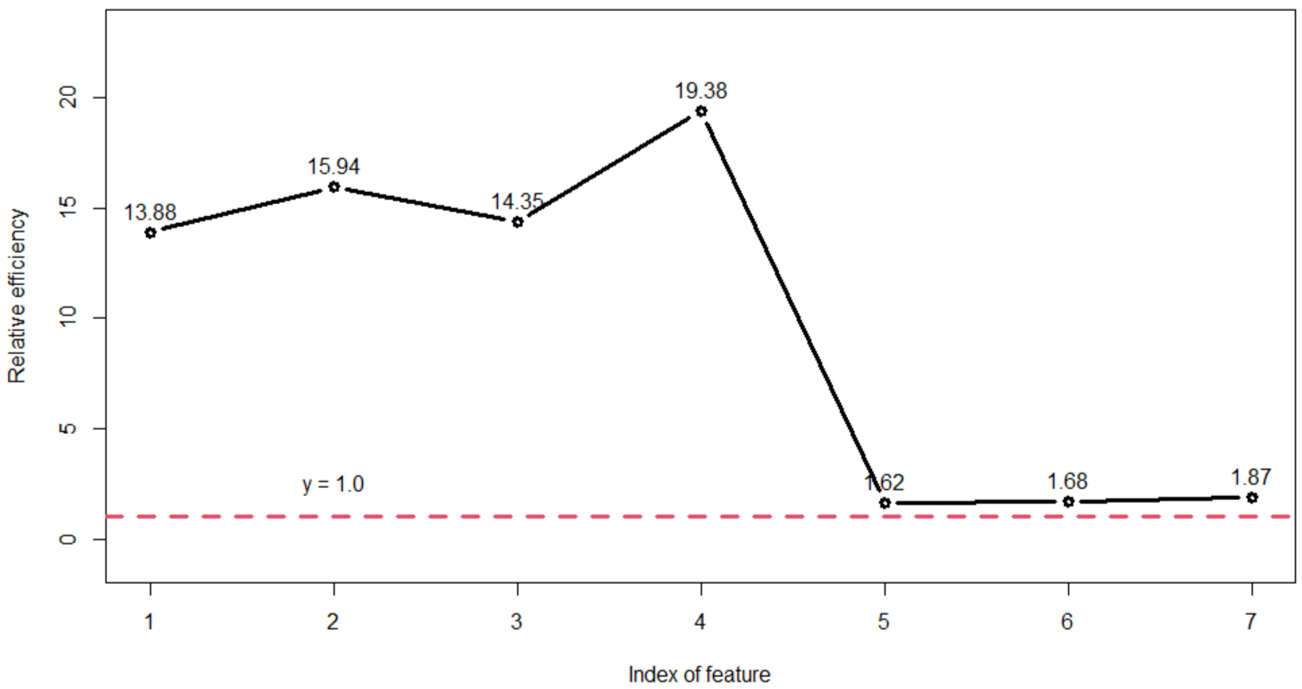}
\caption{Relative efficiency of the proposed method with respect to the baseline method. }
\label{fig:relativeEfficiency}
\end{minipage}
\end{figure}

Fig.~\ref{fig:stdcoefsimu} shows the standard deviation (std) of coefficients estimation from the two methods. It can be seen that the standard deviation of estimation from the proposed method is significantly smaller than those from baseline method in both stages.
To quantify this advantage, the relative efficiency of the proposed method with respect to the baseline method, defined as $e(T_1, T_2) = \frac{var(T_{2})}{var(T_{1})}$, is plotted in Fig. ~\ref{fig:relativeEfficiency}, where $T_1$ and $T_2$ represent the estimators for $\hat{\boldsymbol{\beta}}$ in the proposed method and baseline method, respectively.
It can be seen that the relative efficiency of the estimator in the proposed method for the coefficients in stage 1 is very big, as high as 19.38. And for the estimator for the coefficients in stage 2, the lowest relative efficiency is still $1.62$. This indicates that the efficiency of the parameter estimation is improved $62\% ~ 1838\%$ in the proposed methodology.
The reason for such improvement is straightforward: in the proposed methods, all available data is involved in the parameter estimation, while in the baseline method, only a small subset of the data is involved in later stages.

For the estimation of cutoff points, the baseline method and the proposed method give the mean estimations as $\{-2.201, 2.192,  0.536\}$ and $\{-2.199, 2.197,  0.506\}$, respectively. 
And the standard of deviation of the estimations by the baseline and proposed methods are $\{0.061, 0.064,  0.178\}$ and $\{0.057, 0.062, 0.119\}$, respectively.
Compared with the ground truth, $ \{ \boldsymbol{\theta}\}_{ground ~ truth} =\{-2.2, 2.2, 0.5\}$,
both methods are functional, yet the proposed approach yields a lower mean deviation and standard deviation. 

Overall, the proposed 2-stage model provides a more accurate and  efficient estimation since it involves all data in the parameter estimation.
Also, in implementation, the proposed algorithm enjoys an advantage with regard to computational convenience since we estimated all parameters in $\zeta$ at once. Additionally, the common coefficients assumption in the proposed method reduces the number of parameters to be estimated from $4+7+2+1 = 14$ to $7+2+1 = 10$ since the coefficients for features in stage 1 are not repeatedly estimated in later stages, but instead assumed to be consistent.

\section{Case Study}

In this section,  we analyze a real data set in \cite{zev2021creation} and demonstrate the usefulness of our proposed approach in the sense of being able to learn from data to effectively evaluate gallstone. In particular, the real clinical data is modeled as a 2-stage sequential decision-making model.
We have $N_{1} = 316$ patients in the first stage. Based on the missing profile, we filter the patients who have features in category 2 into the second stage and label them as $Y_{i}^{(1)}=0.5, i = 1,..., N_{2}$. As a result, $N_{2} = 189$.
Note that here we made the assumption that if a patient has available data in the next stage, the patient should be labeled as ``0.5" at the current stage. The reason for such assumption is that we believe that the provider only recommends further diagnostic interventions to a patient who has indeterminate presentation at the current stage (``0.5"). In other words, the real data comes from the provider's knowledge and experience, and the missing pattern is not missing at random but deliberately interfered with.
Note that features in $\boldsymbol{X}^{(1)}$ include hemolytic disease, gender, age, body mass index (BMI), total bilirubin, alanine transaminase (ALT), aspartate transaminase (AST), lipase, amalyse, gamma-glutamyl transferase (GGT), alkaline phosphatase. These features include demographic and laboratory tests. For features in $\boldsymbol{X}^{(2)}$, based on $\boldsymbol{X}^{(1)}$, it involves  imaging data, including CBD diameter and the presence of CBDS by ultrasound (US) and magnetic resonance imaging (MRI) examinations, respectively. See Table \ref{tbl:CBDS} in Appendix C for descriptive statistics.
 \begin{figure}[t]
\centering
        \includegraphics[width=0.8\linewidth]{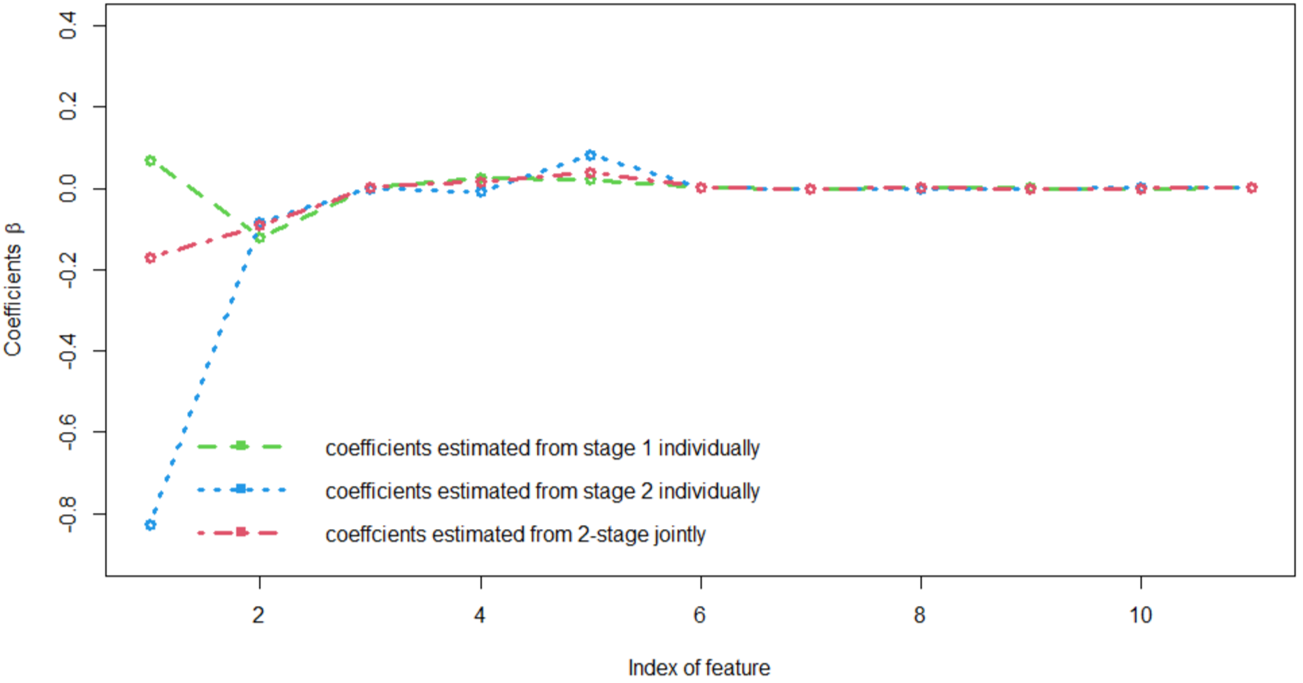}
    \caption{Coefficients comparison between coefficients obtained from 2 methods for features in stages 1.
    }
    \label{fig:coef}
\end{figure}

Fig. ~\ref{fig:coef} visualizes the coefficients for features in category 1 estimated by baseline and proposed approach, respectively.
It can be seen that with the baseline approach, the coefficients estimated by the two individual models are similar for most features, $X_{2} \sim X_{11}$, but significantly different for $X_{1}$.
In the proposed 2-stage model, the coefficients estimated for feature $X_{1} \sim X_{11}$ seem to resemble a weighted average of the coefficients obtained from baseline individual models.
Also, compared with the baseline method, the coefficients estimated by the proposed method have no dramatic spikes. Also, with the common coefficients assumption, newly added features will not affect the weights of existing ones. More importantly, in the proposed method, no matter how many stages a provider utilized, the calculation can be done at once efficiently. 



Our model and method can be useful for providing the treatment recommendation for a new patient, and a diagnostic mobile App was developed in 2021 in Apple Store based on this research. To be concrete, when a new patient presents, given the trained proposed method, the provider (e.g., clinician or doctor), can first recommend the patient undergo diagnostic tests included in category 1, and calculate the $\hat{Y}^{(1)}$ accordingly. Based on the $\hat{L}_1$ and $\hat{U}_1$, the patient will be labeled as healthy (``1"), sick (``0"), or indeterminate and proceed to the next stage (``0.5"). Then, only when the patient's presentation is indeterminate,  additional testing such as ultrasound will be undertaken.
In other words, only the necessary data are actively collected sequentially for sequential decision-making.
This approach simplifies and expedites the diagnostic workup.
Additionally, the provider can adjust the multistage sequential decision-making model flexibly by designing the number of stages and the features in each stage, also see Section 2 for the details on the feature grouping recommendations.   




\section{Conclusion and Discussion}
In this study, to augment healthcare clinical decision-making based on limited examinations, a multistage sequential decision-making model is developed to actively collect only minimal necessary diagnostic data.
To illustrate the proposed method, we optimized a 2-stage model on synthetic data in the simulation study and applied the proposed algorithm to pediatric patients with suspicion for obstructive common bile duct stones in an effort to more efficiently and accurately arrive at a diagnosis that allowed expedited intervention with ERCP.
In both the simulation study and real case study on collected pediatric data, the proposed method is more effective, stable, interpretable, and simpler in estimating the coefficients and cutoff points for all stages.
It estimates all parameters simultaneously  
and fully considers the cumulative characteristics between these stages by making common coefficients assumption.

One limitation of the proposed methodology is that it involves all variables without selection.
A potential future improvement to the model will be the addition of  $L_{1}$ regularization term to the objective function to realize the variable selection.

Also, we can extend our proposed algorithm to satisfy more real-world cases with special requirements. For example, consideration of expense, delay to intervention, and need for invasive testing can be specified and minimized with modification to the objective function as
$\min _{N_{1}, N_{2}, \zeta}-\log \mathcal{L}(\zeta \mid X, Y)+\lambda\left(c_{1} N_{1}+c_{2} N_{2}\right)$,
where the $c_1, c_2$ are the relevant cost per patient in stage 1 and stage 2, respectively. $\lambda$ is the tuning parameter to adjust the weight between the negative likelihood and the relevant cost to the test. Note that here the $N_1, N_2$ can also be variables to be optimized over since the relevant cost depends on the number of patients in each stage.

In healthcare settings, the penalty for misdiagnosis can vary based on the patient and diagnosis. For example, false-positive detection may lead to unnecessary procedures or excess intervention, while false-negative detection may result in missed or delayed treatment. For such cases with weighted penalties on different types of misclassification, we can modify the objective function by specifying the penalties as
$\min _{\zeta}-\log \mathcal{L}(\zeta \mid X, Y)-\lambda_{1} \sum_{i} I\left(\left(y_{i}=0.5 \mid 1\right), \hat{y}_{i}=0\right)-\lambda_{2} \sum_{i} I\left(\left(y_{i}=0 \mid 0.5\right) \& \hat{y}_{i}=1\right)$,
where the $\lambda_1$ and  $\lambda_2$ are the tuning parameter for false-negative prediction and false-positive prediction, respectively. 

The proposed algorithm is also applicable to a broad range of decision-making scenarios outside of healthcare as it utilizes active learning to drive the sequential collection of minimal requisite information, rather than all possible information simultaneously. This will be an area of future development. 


\bigskip

\newpage
\section*{Appendix}
\subsection*{Appendix A -- Alternative Notation for Likelihood and Gradient for 2-Stage Case}
Take our 2-stage illustrating case as an example, the log-likelihood function can be written as:
\begin{equation}
\begin{aligned}
\ell(\boldsymbol{\zeta})= &\sum_{i, n_{i} \in G_{1,1}} \log \left\{1-F\left(U_{1}^{*}-\boldsymbol{\beta}^{(1)^{\top}} \boldsymbol{x}_{i}^{(1)}\right)\right\}+\sum_{i, n_{i} \in G_{1,0}} \log F\left(L_{1}^{*}-\boldsymbol{\beta}^{(1)^{\top}} \boldsymbol{x}_{i}^{(1)}\right)+ \\
 &\sum_{i, n_{i} \in G_{1,0.5}} \log \left\{F\left(U_{1}^{*}-\boldsymbol{\beta}^{(1)\top} \boldsymbol{x}^{(1)}_{i}\right)-F\left(L_{1}^{*}-\boldsymbol{\beta}^{(1)\top} \boldsymbol{x}_{i}^{(1)}\right)\right\}+ \\
 &\sum_{i, n_{i} \in G_{2,1}} \log \left\{1-F\left(C_{2}^{*}-\boldsymbol{\beta}^{(2)\top} \boldsymbol{x}_{i}^{(2)}\right)\right\}+\sum_{i, n_{i} \in G_{2,0}} \log F\left(C_{2}^{*}-\boldsymbol{\beta}^{(2)\top} \boldsymbol{x}_{i}^{(2)}\right)
\end{aligned}
\label{eq:lkhd2stage}
\end{equation}

Since here $\boldsymbol{\beta}^{(2)} = \left\{ \boldsymbol{\beta}^{(1)}, \beta_{1}; \beta_{2}, ...,\beta_{p_2} \right\}$, to avoid any potential confusing notation, we rewrite them as: 
\[ \boldsymbol{\beta}^{(1)} = \{\alpha_1, ...,\alpha_{m}, ..., \alpha_{p1}\}, \quad \boldsymbol{\beta}^{(2)} = \{\alpha_1, ...,\alpha_{m},..., \alpha_{p1}; \beta_{1},..., \beta_{m}, ...,\beta_{p_2} \} \]
\[ \boldsymbol{\zeta} = \{\alpha_1, ...,\alpha_{m},..., \alpha_{p1}; \beta_{1},..., \beta_{m}, ...,\beta_{p_2}; \quad L_{1}^{*}, U_{1}^{*},  C_{2}^{*}  \} \]

Given the joint log-likelihood function, we can write the derivative for each coefficient in $\boldsymbol{\zeta}$ individually accordingly.

For the common coefficients, $\alpha_{m}$ for $m = 1,..., p_1$, we have:
\begin{equation}
\begin{aligned}
\frac{\partial \ell(\boldsymbol{\zeta})} { \partial \alpha_{m} } =
&\sum_{i, \forall n_i \in G_{1,1}}
 \frac{ {x}_{i, m}^{(1)}  f\left(U_1^{*}-\boldsymbol{\beta}^{(1)\top} \boldsymbol{x}_{i}^{(1)}\right) }
{ 1- F\left(U_1^{*}-\boldsymbol{\beta}^{(1)\top} \boldsymbol{x}_{i}^{(1)}\right) } +
\sum_{i, \forall n_i \in G_{1,0}}
 \frac{ -{x}_{i, m}^{(1)}  f\left(L_1^{*}-\boldsymbol{\beta}^{(1)\top} \boldsymbol{x}_{i}^{(1)}\right) }
{ F\left(L_1^{*}-\boldsymbol{\beta}^{(1)\top} \boldsymbol{x}_{i}^{(1)}\right) } +\\
&\sum_{i, \forall n_i \in G_{1,0.5}}
 \frac{ -{x}_{i, m}^{(1)} \left[ f\left(U_1^{*}-\boldsymbol{\beta}^{(1)\top} \boldsymbol{x}_{i}^{(1)}\right) - f\left(L_1^{*}-\boldsymbol{\beta}^{(1)\top} \boldsymbol{x}_{i}^{(1)}\right) \right] }
{  F\left(U_1^{*}-\boldsymbol{\beta}^{(1)\top} \boldsymbol{x}_{i}^{(1)}\right)-F\left(L_1^{*}-\boldsymbol{\beta}^{(1)\top} \boldsymbol{x}_{i}^{(1)}\right) } + \\
&\sum_{i, \forall n_i \in G_{2,1}}
 \frac{ {x}_{i, m}^{(2)}  f\left(C_2^{*}-\boldsymbol{\beta}^{(2)\top} \boldsymbol{x}_{i}^{(2)}\right) }
{ 1- F\left(C_2^{*}-\boldsymbol{\beta}^{(2)\top} \boldsymbol{x}_{i}^{(2)}\right) } +
\sum_{i, \forall n_i \in G_{2,0}}
 \frac{ {x}_{i, m}^{(2)}  f\left(C_2^{*}-\boldsymbol{\beta}^{(2)\top} \boldsymbol{x}_{i}^{(2)}\right) }
{ F\left(C_2^{*}-\boldsymbol{\beta}^{(2)\top} \boldsymbol{x}_{i}^{(2)}\right) }
\label{eq:gradalpha}
\end{aligned}
\end{equation}

For the coefficients for newly added features in stage 2, $\beta_{m}$ for $m = 1,..., p_2$, we have:
\begin{equation}
\begin{array}{c}
\frac{\partial \ell(\boldsymbol{\zeta})} { \partial \beta_{m} } =
\sum_{i, \forall n_i \in G_{2,1}}
 \frac{ {x}_{i, m}^{(2)}  f\left(C_2^{*}-\boldsymbol{\beta}^{(2)\top} \boldsymbol{x}_{i}^{(2)}\right) }
{ 1- F\left(C_2^{*}-\boldsymbol{\beta}^{(2)\top} \boldsymbol{x}_{i}^{(2)}\right) } +
\sum_{i, \forall n_i \in G_{2,0}}
 \frac{ {x}_{i, m}^{(2)}  f\left(C_2^{*}-\boldsymbol{\beta}^{(2)\top} \boldsymbol{x}_{i}^{(2)}\right) }
{ F\left(C_2^{*}-\boldsymbol{\beta}^{(2)\top} \boldsymbol{x}_{i}^{(2)}\right) }
\label{eq:gradbeta2}
\end{array}
\end{equation}
For the cutoff points in all 2 stages, we have:
\begin{equation}
\begin{array}{c}
\frac{\partial \ell(\boldsymbol{\zeta})} { \partial U_1^{*} } =
\sum_{i, \forall n_i \in G_{1,1}}
 \frac{ -f\left(U_1^{*}-\boldsymbol{\beta}^{(1)\top} \boldsymbol{x}_{i}^{(1)}\right) }
{ 1- F\left(U_1^{*}-\boldsymbol{\beta}^{(1)\top} \boldsymbol{x}_{i}^{(1)}\right) } +
\sum_{i, \forall n_i \in G_{1,0.5}}
 \frac{  f\left(U_1^{*}-\boldsymbol{\beta}^{(1)\top} \boldsymbol{x}_{i}^{(1)}\right) }
{  F\left(U_1^{*}-\boldsymbol{\beta}^{(1)\top} \boldsymbol{x}_{i}^{(1)}\right)-F\left(L_1^{*}-\boldsymbol{\beta}^{(1)\top} \boldsymbol{x}_{i}^{(1)}\right) }
\label{eq:gradU}
\end{array}
\end{equation}
\begin{equation}
\begin{array}{c}
\frac{\partial \ell(\boldsymbol{\zeta})} { \partial L_1^{*} } =
\sum_{i, \forall n_i \in G_{1,0}}
 \frac{ f\left(L_1^{*}-\boldsymbol{\beta}^{(1)\top} \boldsymbol{x}_{i}^{(1)}\right) }
{ F\left(L_1^{*}-\boldsymbol{\beta}^{(1)\top} \boldsymbol{x}_{i}^{(1)}\right) } +
\sum_{i, \forall n_i \in G_{1,0.5}}
 \frac{  - f\left(L_1^{*}-\boldsymbol{\beta}^{(1)\top} \boldsymbol{x}_{i}^{(1)}\right) }
{  F\left(U_1^{*}-\boldsymbol{\beta}^{(1)\top} \boldsymbol{x}_{i}^{(1)}\right)-F\left(L_1^{*}-\boldsymbol{\beta}^{(1)\top} \boldsymbol{x}_{i}^{(1)}\right) }
\label{eq:gradL}
\end{array}
\end{equation}
\begin{equation}
\begin{array}{c}
\frac{\partial \ell(\boldsymbol{\zeta})} { \partial C_2^{*} } =
\sum_{i, \forall n_i \in G_{2,1}}
 \frac{ -f\left(C_2^{*}-\boldsymbol{\beta}^{(2)\top} \boldsymbol{x}_{i}^{(2)}\right) }
{ 1- F\left(C_2^{*}-\boldsymbol{\beta}^{(2)\top} \boldsymbol{x}_{i}^{(2)}\right) } +
\sum_{i, \forall n_i \in G_{2,0.5}}
 \frac{  f\left(C_2^{*}-\boldsymbol{\beta}^{(2)\top} \boldsymbol{x}_{i}^{(2)}\right) }
{  F\left(C_2^{*}-\boldsymbol{\beta}^{(2)\top} \boldsymbol{x}_{i}^{(2)}\right) }
\label{eq:gradC}
\end{array}
\end{equation}

\subsection*{Appendix B -- Mean Prediction Metrics Comparison of the Baseline Method and Proposed Method in Simulation Study}

\begin{figure}[htb!]
    \centering
    \includegraphics[width =\linewidth]{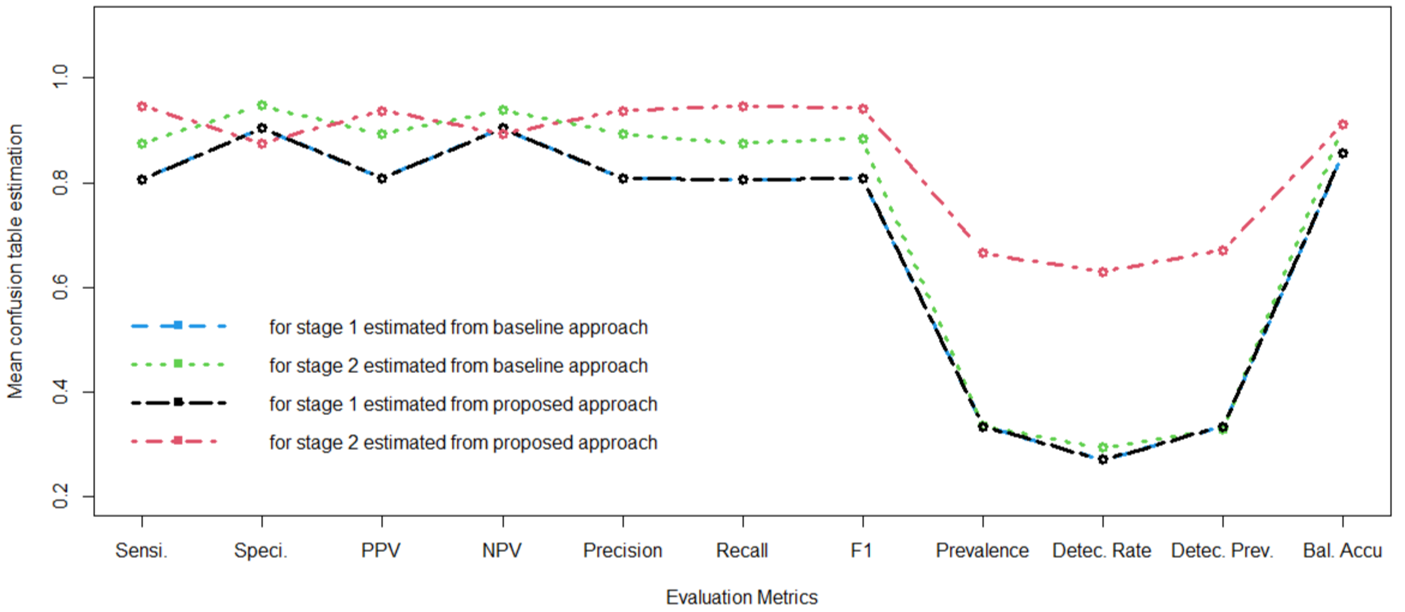}
    \caption{Mean prediction metrics for 2 stages with coefficients estimated from baseline approach individually (blue and green line), and coefficients estimated from proposed method jointly (red and black line). Note that the blue and black lines are too close to each other to distinguish them from each other visually.}
    \label{fig:CFtablesimu}
\end{figure}

Fig. ~\ref{fig:CFtablesimu} shows the mean prediction metrics for prediction for 2 stages with coefficients estimated from the two methods respectively. The evaluation metrics include sensitivity (Sensi.), specificity (Speci.), positive predictive values (PPV), negative predictive values (NPV), and balanced accuracy, precision, recall, F1 value, prevalence, detection rate (Detec. Rate), detection prevalence (Detec. Prev.), and balanced accuracy (Bal. Accu).
From the predictive performance comparison in Fig. ~\ref{fig:CFtablesimu}, it can be seen that the proposed method has a significant advantage for the prediction for stage 2 with almost higher values in every index, especially the prevalence, and detection rate, and detection prevalence. And for the prediction for stage 1, there's no significant difference between the proposed method and baseline method, as we can see that the blue and black lines are too close to each other to distinguish them from each other visually.

\subsection*{Appendix C -- Table 1. Descriptive Statistics for Demographic, Laboratory, and Imaging Data in Case Study}
\begin{table}[htb!]
\centering
\caption{Descriptive statistics for demographic, laboratory, and imaging data}
\label{tbl:CBDS}
\resizebox{\textwidth}{!}{%

\begin{tabular}{llll}
\hline
& CBD Stone Present & No CBD Stone Present & p-value \\
\hline Total patients & 120 & 196 & \\
IOC & 81 & 196 & \\
ERCP & 39 & 0 & \\
\hline Hemolytic disease & $23(19.2 \%)$ & $49(25.0 \%)$ & $0.270$ \\
Female & $81(67.5 \%)$ & $123(62.8 \%)$ & $0.400$ \\
Age $(\mathrm{yrs})$ & $13.8 \pm 3.3$ & $13.6 \pm 3.6$ & $0.883$ \\
BMI $\left(\mathrm{kg} / \mathrm{m}^{2}\right)$ & $25.7 \pm 8.0$ & $24.5 \pm 8.7$ & $0.125$ \\
\hline
Total bilirubin $(\mathrm{mg} / \mathrm{dL})$ & $5.6 \pm 8.3$ & $2.6 \pm 5.2$ & $<0.001$ \\
ALT (U/L) & $258.5 \pm 195.8$ & $132.9 \pm 146.2$ & $<0.001$ \\
AST (U/L) & $153.3 \pm 135.6$ & $87.9 \pm 110.6$ & $<0.001$ \\
Lipase (U/L) & $578.4 \pm 1516.9$ & $268.4 \pm 517.8$ & $0.538$ \\
Amylase (U/L) & $113.6 \pm 197.8$ & $119.3 \pm 190.9$ & $0.011$ \\
GGT (U/L) & $296.0 \pm 202.0$ & $232.5 \pm 216.1$ & $0.024$ \\
Alkaline phosphatase (U/L) & $247.1 \pm 189.6$ & $179.7 \pm 98.1$ & $0.014$ \\
\hline
CBD diameter (mm, by US) & $7.4 \pm 3.8$ & $5.6 \pm 3.8$ & $<0.001$ \\
Presence of CBD stone (US) & $11(13.4 \%)$ & $3(3.1 \%)$ & $0.012$\\
CBD diameter (mm, by MRI) & $9.3 \pm 3.8$ & $8.3 \pm 2.6$ & $0.420$ \\
Presence of CBD stone (MRI) & $28(54.9 \%)$ & $10(19.6 \%)$ & $0.041$\\
\hline
\end{tabular}
}
\end{table}

\end{document}